\newif\ifRAL
\RALfalse 	

\newif\ifTR
\TRfalse 	

\newif\ifPrePrint
\PrePrinttrue

\newif\ifDraft
\Draftfalse

\ifRAL
\documentclass[letterpaper, 10 pt, journal, twoside]{IEEEtran}
\else
\documentclass[letterpaper, 10 pt, conference]{ieeeconf}
\fi

\IEEEoverridecommandlockouts

\ifRAL
\else
	\def\baselinestretch{0.960}
\fi		

\makeatletter

\let\proof\@undefined
\let\endproof\@undefined
\makeatother

\usepackage{amsmath} 
\usepackage{amssymb}  
\usepackage{amsthm}
\usepackage{amsfonts}
\usepackage{mathtools}

\usepackage{bm}
\providecommand{\bm}{\pmb}

\theoremstyle{definition}

\theoremstyle{remark}

\usepackage{verbatim}

\usepackage{color}
\usepackage{subcaption}
\usepackage{graphicx}
\usepackage{caption}

\usepackage{optidef}

\usepackage{times}

\usepackage{xspace}
\usepackage{siunitx}

\usepackage[ruled,vlined,linesnumbered]{algorithm2e}
\usepackage[noend]{algorithmic}

\graphicspath{{images/}}

\usepackage[us]{datetime}
\usepackage{caption}
\usepackage[usenames,dvipsnames,table]{xcolor}

\usepackage{hyperref} 
\hypersetup{
    colorlinks,
    citecolor=black,
    filecolor=black,
    linkcolor=black,
    urlcolor=black,
    pdfauthor={},
    pdfsubject={},
    pdftitle={}
}

\usepackage[capitalise]{cleveref}
\crefname{equation}{}{}

\usepackage{todonotes}

\usepackage{graphicx}

\usepackage{latexsym}
\usepackage{color}

\usepackage{cite}

\usepackage{caption}

\usepackage{tikz}
\usetikzlibrary{shapes,arrows}

\usepackage{tabularx, booktabs}
\newcolumntype{Y}{>{\centering\arraybackslash}X}
\usepackage{multirow}
\usepackage{paralist}
\usepackage{units}

\usepackage[final]{changes}

\usepackage{acro}

\DeclareAcronym{ASL}{short = ASL, long = Autonomous Systems Lab}
\DeclareAcronym{OMAV}{short = OMAV, long = Omnidirectional Micro Aerial Vehicle}
\DeclareAcronym{MAV}{short = MAV, long = Micro Aerial Vehicle}
\DeclareAcronym{DOF}{short = DOF, long = degrees of freedom}
\DeclareAcronym{PBC}{short = PBC, long = passivity-based control}
\DeclareAcronym{PH}{short = PH, long = Port-Hamiltonian}
\DeclareAcronym{NDT}{short = NDT, long = non-destructive testing}
\DeclareAcronym{PEMS}{short = PEMS, long = Power and Energy Monitoring System}
\DeclareAcronym{WTC}{short = WTC, long = wrench tracking controller}
\DeclareAcronym{MBE}{short = MBE, long = momentum-based wrench estimator}
\DeclareAcronym{ASIC}{short = ASIC, long = Axis-Selective Impedance Control}
\DeclareAcronym{MPC}{short = MPC, long = Model Predictive Control}
\DeclareAcronym{APhI}{short = APhI, long = Aerial Physical Interaction}
\DeclareAcronym{FT}{short = FT, long = Force-Torque}

\newcommand{\tolie}[1]{\left[{#1}\right]_\times}

\renewcommand{\vec}[1]{\bm{#1}}		
\newcommand{\matr}[1]{\bm{#1}}		
\newcommand{\blkdiag}[1]{\text{blkdiag}\left(#1\right)}		
\newcommand{\mat}[1]{\bm{#1}}		
\newcommand{\diag}[1]{\text{diag}\left(#1\right)}		
\newcommand{\nR}[1]{\mathbb{R}^{#1}}		
\newcommand{\SO}[1]{\mathrm{SO}(#1)}		
\renewcommand{\matrix}[1]{\begin{bmatrix} #1 \end{bmatrix}}	
\newcommand{\upperRomannumeral}[1]{\uppercase\expandafter{\romannumeral#1}}	

\newcommand{\transpose}{^\top}

\renewcommand{\frame}[1]{\mathcal{F}_{#1}}		

\newcommand{\pos}{\vec{r}}				
\newcommand{\vel}{\vec{v}}				

\newcommand{\eye}[1]{\matr{I}_{#1}}
\newcommand{\trace}[1]{\text{tr}\left(#1\right)}
\newcommand{\zeros}[1]{\matr{0}_{#1}}

\newcommand{\angVel}{\vec{\omega}}

\newcommand{\momentum}{\bm{p}}    
\newcommand{\momentumEst}{\hat{\bm{p}}}
\newcommand{\momentumEstDot}{\dot{\hat{\bm{p}}}}
\newcommand{\wrench}{\bm{w}}

\newcommand{\posRef}{\pos_d}

\newcommand{\twist}{\bm{t}}    

\newcommand{\inertia}{\bm{J}}

\newcommand{\totalInertia}{\bm{M}}
\newcommand{\totalInertiaDesired}{\bm{M}_d}
\newcommand{\mass}{m}

\newcommand{\tankState}{x_t}
\newcommand{\tankStatedot}{\dot{x}_t}
\newcommand{\tankInput}{u_t}
\newcommand{\tankEnergy}{\hamiltonianTank}

\newcommand{\tankEps}{\epsilon_t}
\newcommand{\tankLowerLimit}{\mathcal{H}_t^-}
\newcommand{\tankUpperLimit}{\mathcal{H}_t^+}

\newcommand{\RB}{\bm{R}}
\newcommand{\Rdes}{\bm{R}_d}

\newcommand{\hamiltonianKin}{\mathcal{H}_{kin}}

\newcommand{\hamiltonianSpring}{\mathcal{H}_{spr}}

\newcommand{\hamiltonianCL}{\mathcal{H}_{cl}}
\newcommand{\hamiltonianCLdot}{\dot{\mathcal{H}}_{cl}}
\newcommand{\hamiltonianCLtotaldot}{\dot{\bar{\mathcal{H}}}_{cl}}

\newcommand{\hamiltonianObserver}{\mathcal{H}_{obs}}

\newcommand{\hamiltonianTank}{\mathcal{H}_{t}}
\newcommand{\hamiltonianTankdot}{\dot{\mathcal{H}}_{t}}

\newcommand{\Kspring}{\mat{K}_p}
\newcommand{\Kspringlin}{\mat{K}_{p,lin}}
\newcommand{\Kspringrot}{\mat{K}_{p,rot}}
\newcommand{\Kspringbar}{\bar{\mat{K}}_p}
\newcommand{\Kdamping}{\mat{K}_d}
\newcommand{\Kdampinglin}{\mat{K}_{d,lin}}
\newcommand{\Kdampingrot}{\mat{K}_{d,rot}}
\newcommand{\Kdampingbar}{\bar{\mat{K}}_d}

\newcommand{\Ktracking}{\bm{K}_{p,tr}}
\newcommand{\Kintegral}{\bm{K}_{i,tr}}
\newcommand{\Kobserver}{\bm{K}_o}

\newcommand{\wrenchInt}{\wrench_{int}}
\newcommand{\wrenchIntEst}{\hat{\wrench}_{int}}

\newcommand{\wrenchExtEst}{\hat{\wrench}_{ext}}

\newcommand{\wrenchExtEstDot}{\dot{\hat{\wrench}}_{ext}}
\newcommand{\wrenchIntDes}{\wrench_{int,d}}
\newcommand{\wrenchIntErr}{\wrench_{int,e}}

\newcommand{\wrenchGravity}{\wrench_{g}}

\newcommand{\wrenchCommand}{\wrench_{c}}
\newcommand{\wrenchExt}{\wrench_{ext}}
\newcommand{\wrenchDist}{\wrench_{dist}}

\newcommand{\wrenchCmd}{\wrench_{c}}
\newcommand{\wrenchCmdImpedance}{\wrench_{c,imp}}
\newcommand{\wrenchCmdTracking}{\wrench_{c,tr}}

\newcommand{\dampingone}{d_1}
\newcommand{\dampingtwo}{d_2}
\newcommand{\drainingA}{\bm{\omega}_1}

\newcommand{\drainingBC}{\bm{\omega}_{2}}
\newcommand{\drainingDE}{\bm{\omega}_{3}}

\newcommand{\error}{\bm{e}}
\newcommand{\errorPos}{\bm{e}_{lin}}
\newcommand{\errorAng}{\bm{e}_{rot}}
\newcommand{\Cmatrix}{\bm{C}}
\newcommand{\Mmatrix}{\bar{\bm{M}}}

\newcommand{\valvesB}{\gamma_{2,lin}}
\newcommand{\valvesC}{\gamma_{2,ang}}
\newcommand{\valvesD}{\gamma_{3,lin}}
\newcommand{\valvesE}{\gamma_{3,ang}}
\newcommand{\powerA}{p_1}
\newcommand{\powerBC}{p_{2}}

\newcommand{\powerDE}{p_{3}}

\newcommand{\powerMax}{p_{tot}^+}

\newcommand{\GammaA}{\vec{\Gamma}_1}
\newcommand{\GammaBC}{\vec{\Gamma}_2}
\newcommand{\GammaDE}{\vec{\Gamma}_3}

\newcommand{\effortA}{\vec{y}_1}
\newcommand{\effortBC}{\vec{y}_2}
\newcommand{\effortDE}{\vec{y}_3}

\ifRAL 
\author{Maximilian Brunner, Livio Giacomini, Roland Siegwart, and Marco Tognon
	
	\thanks{Manuscript received: DD,\,MM,\,YY; Revised DD,\,MM,\,YY ; Accepted  DD,\,MM,\,YY.}
	\thanks{This paper was recommended for publication by [Editor] upon evaluation of the Associate Editor and Reviewers' comments. 
	This work was partially funded by ...} 
	
	\thanks{$^1$ Affiliation, {\tt \footnotesize \href{mailto:first.author@xx.xx}{first.author@xx.xx}, \href{mailto:last.author@xx.xx}{first.author@xx.xx}}
	}
	
	\thanks{$^2$ Affiliation {\tt \footnotesize \href{mailto:last.author@xx.xx}{first.author@xx.xx}}
	}

	\thanks{Digital Object Identifier (DOI): see top of this page.}	
}

\else 
\author{Maximilian Brunner, Livio Giacomini, Roland Siegwart, and Marco Tognon
	\thanks{All authors are with the Autonomous Systems Lab at ETH Zurich. Email: {\tt \footnotesize \href{mailto:maximilian.brunner@mavt.ethz.ch}{maximilian.brunner@mavt.ethz.ch}}
	}
	\thanks{This research was partially supported by the National  Center  of  Competence  in Research (NCCR)  Digital Fabrication, the NCCR Robotics, and the Armasuisse Science and Technology.}
}
\fi

\ifRAL 
\title{Energy Tank-Based Policies for Robust Aerial Physical Interaction with Moving Objects}

\else 
\title{Energy Tank-Based Policies for Robust Aerial Physical Interaction with Moving Objects}
\fi

\ifPrePrint
\makeatletter
\def\ps@titlepagestyle{
	\def\@oddfoot{}\def\@evenfoot{}
	\def\@oddhead{\textcolor{red}{\sf\footnotesize Preprint version, final version at http://ieeexplore.ieee.org/ \hfill IEEE International Conference on Robotics and Automation 2022}}
	\def\@evenhead{\textcolor{red}{\sf\footnotesize  Preprint version, final version at http://ieeexplore.ieee.org/  \hfill IEEE International Conference on Robotics and Automation 2022}}%
}%
\makeatother
\makeatletter
\def\ps@headings{
	\def\@oddfoot{\textcolor{red}{\sf\footnotesize  Preprint version, final version at http://ieeexplore.ieee.org/ \hfill \thepage \;\;~\hfill~\hfill IEEE International Conference on Robotics and Automation 2022}}\def\@evenfoot{\hfill\thepage\hfill}
	\def\@oddhead{}\def\@evenhead{}%
}%
\makeatother
\fi	

\ifDraft
\makeatletter
\def\ps@titlepagestyle{
	\def\@oddfoot{}\def\@evenfoot{}
	\def\@oddhead{\textcolor{red}{\sf Draft version  \hfill Confidential}}
	\def\@evenhead{\textcolor{red}{\sf  Draft version  \hfill Confidential}}%
}%
\makeatother
\makeatletter
\def\ps@headings{
	\def\@oddfoot{\textcolor{red}{\sf  Draft version  \hfill Confidential}}\def\@evenfoot{\hfill\thepage\hfill}
	\def\@oddhead{}\def\@evenhead{}%
}%
\makeatother
\usepackage{draftwatermark}
\SetWatermarkText{Confidential draft}
\SetWatermarkScale{0.6}
\SetWatermarkLightness{0.9} 

\fi	

\begin{document}

\maketitle

\begin{abstract}
Although manipulation capabilities of aerial robots greatly improved in the last decade, only few works addressed the problem of aerial physical interaction with dynamic environments, proposing strongly model-based approaches. 
However, in real scenarios, modeling the environment with high accuracy is often impossible.
In this work, we aim at developing a control framework for \acp{OMAV} for reliable physical interaction tasks with articulated and movable objects in the presence of possibly unforeseen disturbances, and without relying on an accurate model of the environment. 
Inspired by previous applications of energy-based controllers for physical interaction,  we propose a passivity-based impedance and wrench tracking controller in combination with a momentum-based wrench estimator.
This is combined with an energy-tank framework to guarantee the stability of the system, while energy and power flow-based adaptation policies are deployed to enable safe interaction with any type of passive environment.
The control framework provides formal guarantees of stability, which is validated in practice considering the challenging task of pushing a cart of unknown mass, moving on a surface of unknown friction, as well as subjected to unknown disturbances.
For this scenario, we present, evaluate and discuss three different policies.
\end{abstract}

\ifRAL 
\else 
{} 
\fi

\section{INTRODUCTION}\label{sec:intro}

\ifRAL
\IEEEPARstart{I}{ntro}
\else

\fi
In the last decade, there has been a growing attention to the field of \ac{APhI}~\cite{2021g-OllTogSuaLeeFra}. 
In an effort towards enhancing manipulation capabilities of aerial robots, the problem has been addressed from different aspects, including the design of new control methods and new platforms more suited for interaction tasks.
Those solutions lead to a new generation of aerial manipulators based on fully actuated, multi-, and omnidirectional thrust vehicles~\cite{2021f-HamUsaSabStaTogFra}, capable to generate forces and torques in 6 \ac{DOF}, and equipped with interaction tools like rigid rods~\cite{Ryll2019}, and articulated arms~\cite{2021-BodTogSie}.

However, performing aerial interaction is inherently difficult due to the change of the system dynamics during interaction. 
Generally, we can identify two types of physical interaction: 
\begin{inparaenum}[i)]
	\item the one with a \emph{static} environment, and 
	\item the one with a \emph{dynamic} environment.
\end{inparaenum}

So far the research community mostly addressed \ac{APhI} with static environments where the objective is to maintain contact or deliver a specific interaction force between the aerial robot and a rigid structure, while possibly sliding along its surface.
In this case the environment is passive and its state does not change.  %
Most works rely on position controllers driving the platform equipped with a mechanically compliant tool onto a surface~\cite{2019e-TogTelGasSabBicMalLanSanRevCorFra}, or on impedance controllers~\cite{2018-SuaHerOll}, possibly also in combination with a force-tracking controller~\cite{Meng2019, Bodie2020}. 
Model-based solutions have been presented as well for push-and-slide operations~\cite{Tzoumanikas2020,Nava2020}.

\begin{figure}[t!]
    \centering
    \includegraphics[width=\linewidth]{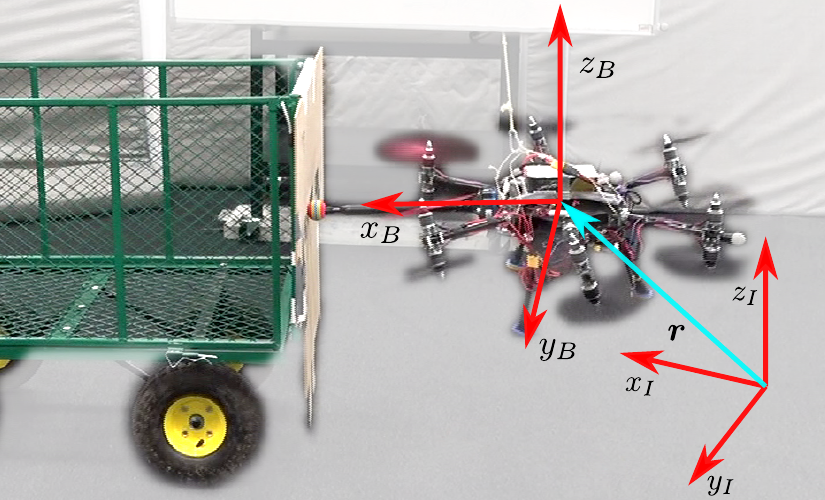}
    \caption{An OMAV pushing a movable cart.}
    \vspace*{-0.3cm}
    \label{fig:frames}
\end{figure}

On the other hand, the less investigated interaction with dynamic environments (e.g. the aerial robot pushing a movable cart as illustrated in \cref{fig:frames}) implies additional challenges related to the change of state of the environment under the robot action, and vice versa. 
If the dynamics of the robot and the environment are perfectly known, a dynamic interaction task can theoretically be executed through a combination of accurate trajectory planning, state estimation of the entire system (robot and environment), and precise hybrid position-force control~\cite{2009-SicSciVilOri}.
Along this line, recent works proposed both model-based and traditional {motion-planning} approaches. 
In \cite{Lee2020} a \ac{MPC} framework has been presented to open a hinged door, while in \cite{Lee2021} the task of pushing a cart has been approached by dynamically updating the aerial robot position reference.
While these approaches are able to tackle interaction tasks with dynamic environments in very structured conditions, no guarantees about their safety nor their robustness against model uncertainties and external disturbances can be made. 
In fact, in real scenarios, especially for aerial robotic applications, the state and physical characteristics of the environment are mostly unknown and unexpected disturbances may appear, drastically degrading performance eventually leading to the instability of the system.

To address the robustness issue during physical interaction, recent works proposed to study the system from an energetic perspective.
In \cite{Stramigioli}, the interaction is modeled employing the concept of power ports, considering the energy flow between interacting subsystems.
When treating the system as an energy exchanging device, it is natural to use passivity as a stability criterion~\cite{1991-SloLi_}.
This principle has proven to be a solid mathematical background in many areas such as in bilateral teleoperation~\cite{Nuno2011}, multi-robot coordination~\cite{Chopra2006,2018h-TogGabPalFra}, and physical human-robot interaction~\cite{tognon2021physical}.
However, many control approaches and actions are not passive by design.  
Therefore, energy storage elements, also called \emph{energy tanks}, were introduced in \cite{Duindam2004}.
They allow to make non-passive actions passive using the stored energy s.t. the total internal energy of the system does not increase~\cite{Secchi2006,Dietrich2017}.
Recently, this energy tank-based approach has been transferred to contact-based aerial inspection, where a similar setup was employed~\cite{Rashad2019Tank}.
However, only interaction with static environments was investigated.
On the contrary, we want to fully exploit the potential of passivity-based control methods to extend aerial physical interaction to dynamic environments ensuring stability and safety.

Using energy-based control techniques, additional safety features can be naturally deployed observing the energy exchange between sub-systems.
In particular, the tank in- and outflow of energy, as well as the energy within the tank itself, can be exploited to implement high-level safety features.
The work in \cite{Shahriari2018} introduced the concept of \textit{power valves} to limit the power exchange between the tank and the system for a ground manipulator.

In this work, we take inspiration from the works of \cite{Rashad2019Tank} and \cite{Shahriari2018} to develop a passivity-based control framework for robust \ac{APhI} with dynamic environments, which can formally guarantee stability and safety.

To this end, we design an impedance controller combined with a wrench tracking PI controller using a momentum-based wrench estimator considering an \ac{OMAV}.
We use an energy tank to restore the system's passivity and to guarantee stable interaction with any passive environment. 
The method does not require the knowledge of environment dynamics which can be time varying, as long as they remain passive.

While the energy tank ensures the overall systems passivity, we employ safety policies that are based on the energy stored within the tank, as well as on the magnitudes of the power flows between the tank and the system. 
These policies can adapt the control inputs to execute the interaction task while bounding the power flows to prevent potentially unsafe behaviors.
To show the stability and safety properties of the proposed control framework, we challenge it with the complex task of pushing a cart of unknown mass under the presence of external disturbances and unknown friction conditions.
In real experiments, we show that a standard interaction control method cannot provide enough stability and safety when physically interacting with dynamic environments under strong uncertainties and external disturbances. 
On the contrary, the introduced passivity-based method allows to meet this fundamental requirement.
Additional practical considerations on different safety policies are provided as well.

\section{MODELING}\label{sec:modeling}
We use two frames for the derivation of the system dynamics: the inertial frame $\frame{I} = \{O_I, \vec{x}_I, \vec{y}_I, \vec{z}_I\}$ and the body-fixed frame $\frame{B} = \{O_B, \vec{x}_B, \vec{y}_B, \vec{z}_B\}$, where $O_*$ represents the origin and $\vec{x}_*,\vec{y}_*,\vec{z}_*$ the primary axes of the frame.
Let $\RB\in \SO{3}$ be the rotation matrix representing the rotation of $\frame{B}$ w.r.t. $\frame{I}$, and let $\pos\in\nR{3}$ be the body position vector, given in $\frame{I}$.
We define the body twist as the stacked angular and linear velocities, $\angVel \in \nR{3}$ and $\vel \in \nR{3}$, respectively, both expressed in $\frame{B}$:
\begin{align}
\twist=\matrix{\angVel \\ \vel}\in\nR{6}.
\end{align}
Then we define the momentum vector $\momentum \in \nR{6}$ as 
\begin{align}
\momentum=\totalInertia\twist,
\end{align}
where $\totalInertia=\blkdiag{\inertia,m\eye{3}}\in\nR{6\times 6}$ represents the generalized inertia tensor, containing the moment of inertia $\inertia \in \nR{3 \times 3}$ and the system mass $\mass \in \nR{}$.
Following the Newton-Euler approach we write the system dynamics as follows:
\begin{multline}
\underbrace{\matrix{ \inertia & \bm{0} \\ \bm{0} & \mass\eye{3}}}_{\totalInertia}
\underbrace{\matrix{\dot\angVel \\ \dot\vel}}_{\dot\twist}=\underbrace{\matrix{ \inertia\tolie{\angVel} & \bm{0} \\ \bm{0} &-\mass\tolie{\angVel}}}_{\Cmatrix}\matrix{\angVel\\ \vel}\\+{\wrenchCommand}+{\wrenchGravity} + {\wrenchExt},\label{eq:dynamics}
\end{multline}
where\footnote{The symbol $\tolie{\cdot}:\nR{3} \rightarrow \mathfrak{so}{(3)}$ represents the symmetric-skew operator such that, given two vector $\vec{y},\vec{x}\in\nR{3}$, $\tolie{\vec{y}}\vec{x}=\vec{y} \times \vec{x}$.} $\wrenchExt=\wrenchDist+\wrenchInt$ comprises both disturbance and interaction wrenches defined by $\wrenchDist \in \nR{6}$ and $\wrenchInt \in \nR{6}$, respectively; $\wrenchGravity \in \nR{6}$ represents the gravity force; and $\wrenchCommand \in \nR{6}$ the commanded wrench produced by the actuators.

\section{INTERACTION CONTROL}\label{sec:control}
Considering an \ac{OMAV} in physical interaction, we employ an interaction control framework composed of an impedance controller combined with a wrench tracking PI controller using a momentum-based wrench estimator.
In this section, we report the main derivations and the relative analysis of stability.
As shown in~\cite{Rashad2019Tank}, it turns out that such controller cannot guarantee the passivity of the system.
In the following sections we shall then show how to guarantee passivity and add an extra safety layer.
  
\subsection{Axis-Selective Impedance controller}
Similar to \cite{Bodie2020a} we introduce an \ac{ASIC}. This gives us the advantage to set the desired virtual mass and inertia of the platform for each axis individually, combined in the desired generalized inertia tensor, $\totalInertiaDesired\in\nR{6 \times 6}$.

We employ a momentum-based wrench observer to estimate the total external wrench acting on the platform, given by \replaced{$\wrenchExtEst=\matrix{\hat{\vec{f}}_{ext}\transpose & \hat{\vec{\tau}}_{ext}\transpose}\transpose \in \nR{6}$}{$\wrenchExtEst\in \nR{6}$}.

The impedance control law is then given by
\begin{align}
\begin{split}
\wrenchCmdImpedance &= \underbrace{\left(\totalInertia\totalInertiaDesired^{-1}-\eye{6}\right)}_{\Mmatrix}\wrenchExtEst-\totalInertia\totalInertiaDesired^{-1}\left(\Kdamping\twist+\Kspring\error\right)\\
&\qquad-\Cmatrix\twist-\wrenchGravity\\
&= \Mmatrix\wrenchExtEst-\left(\Kdampingbar\twist+\Kspringbar\error\right)-\Cmatrix\twist-\wrenchGravity,\label{eq:impedanceCommand}
\end{split}
\end{align}
with $\Kdamping,\Kspring\in\nR{6\times 6}_{>0}$, $\Kdampingbar=\totalInertia\totalInertiaDesired^{-1}\Kdamping$ and $\Kspringbar=\totalInertia\totalInertiaDesired^{-1}\Kspring$.
The error vector $\error=\matrix{\errorAng\transpose & \errorPos\transpose}\transpose\in\nR{6}$ is composed of the angular and linear error, represented by
\begin{subequations}
\begin{align}
\errorAng&=\frac{1}{2}\left(\Rdes\transpose\RB-\RB\transpose\Rdes\right)^\vee\\
\errorPos &=\pos-\posRef,
\end{align}
\end{subequations}
where\footnote{The \textit{vee}-map $(\cdot)^\vee:\mathfrak{so}{(3)}\rightarrow\nR{3}$ is the inverse of the skew-symmetric operator $\tolie{\cdot}:\nR{3}\rightarrow\mathfrak{so}{(3)}$.} $\Rdes\in\SO{3}$ represents the reference attitude and $\posRef\in\nR{3}$ the reference position of the platform. 
The gain matrices are composed from their linear and angular parts, i.e. $\Kspring=\blkdiag{\Kspringrot,\Kspringlin}$ and $\Kdamping=\blkdiag{\Kdampingrot,\Kdampinglin}$, with $\Kspringrot$, $\Kspringlin$, $\Kdampingrot$, $\Kdampinglin\in\nR{3\times 3}$.
\added{
Combining \cref{eq:dynamics}, \cref{eq:impedanceCommand}, and assuming $\wrenchExt=\wrenchExtEst$, we obtain the closed loop dynamics:
\begin{align}
    \totalInertiaDesired\dot\twist=-\Kdamping\twist-\Kspring\error+\wrenchExt.
\end{align}
}

\subsection{Momentum based observer}
The momentum based observer dynamics are~\cite{Ryll2017}:
\begin{align}
\wrenchExtEst&=\Kobserver(\momentum-\momentumEst)\\
\momentumEstDot&=\Cmatrix\twist+\wrenchGravity+\wrenchCmd+\Kobserver(\momentum-\momentumEst),
\end{align}
where $\Kobserver\in\nR{6\times 6}$ is the observer gain matrix.
It follows that the wrench estimate tracks the true external wrench through a first order lowpass filter:
\begin{align}
\wrenchExtEstDot=\Kobserver(\wrenchExt-\wrenchExtEst).
\end{align}

\subsection{Wrench tracking controller}
In addition to the impedance controller we introduce a wrench tracking controller based on a PI law:
\begin{align}
\wrenchCmdTracking&=\Ktracking\wrenchIntErr+\Kintegral\int\wrenchIntErr dt\\
\wrenchIntErr&=\wrenchIntEst-\wrenchIntDes,
\end{align}
with $\Ktracking,\Kintegral\in\nR{6\times 6}$ being the proportional and integral controller gains, respectively.
\added{The estimated interaction wrench is composed of the external force estimate and zero torque, i.e. $\wrenchIntEst=\matrix{\hat{\vec{f}}_{ext}\transpose & \zeros{3}}\transpose$.
}

We can then write the total control command $\wrenchCmd$ as
\begin{align}
\wrenchCmd=\wrenchCmdImpedance+\wrenchCmdTracking.\label{eq:controlCommands}
\end{align}
\begin{figure}[t]
    \centering
    \hspace*{-0.5cm}\includegraphics{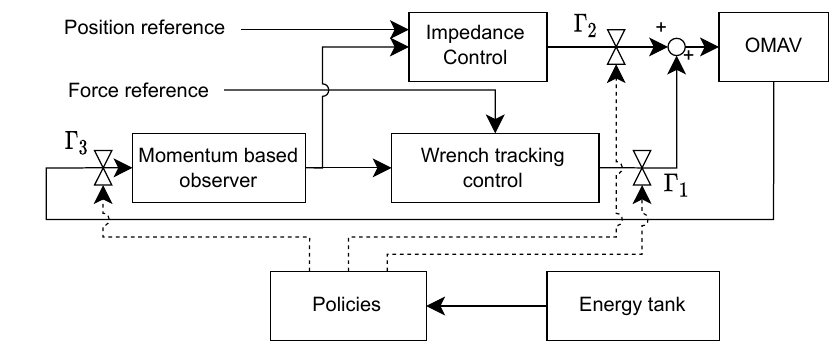}
    \caption{Control block diagram including the valves which can scale the individual inputs.}
    \vspace*{-0.3cm}
    \label{fig:control_block_diagram}
\end{figure}
\Cref{fig:control_block_diagram} shows the control block diagram of the combined impedance and \replaced{wrench}{force} tracking controller.
\subsection{Passivity analysis}
We now analyze the stability of the system. 
In particular, we verify the passivity property considering the closed-loop system energy as storage function, $\hamiltonianCL$.
The latter is defined as the sum of the kinetic energy of the platform $\hamiltonianKin$, the spring energy given by the impedance controller $\hamiltonianSpring$, and the observer energy $\hamiltonianObserver$: 
\begin{align}
\hamiltonianCL = \hamiltonianKin+\hamiltonianSpring+\hamiltonianObserver,
\end{align}
where
\begin{subequations}
\begin{align}
\hamiltonianKin &= \frac{1}{2}\momentum\transpose\totalInertia^{-1}\momentum=\frac{1}{2}\twist\transpose\totalInertia\twist\\
\hamiltonianSpring&=\frac{1}{2}\errorPos\transpose\Kspringlin\errorPos+\frac{1}{2}\trace{\Kspringrot(\eye{3}-\Rdes\transpose\RB)} \\
\hamiltonianObserver&=\frac{1}{2}\momentumEst\transpose\Kobserver\momentumEst.
\end{align}
\end{subequations}
By definition, the closed-loop system is said to be passive w.r.t. the input-output pair $(\twist,\wrenchExt)$ (the power port which acts between the platform and its environment), if the following inequality holds~\cite{1991-SloLi_}:
\begin{align}
\hamiltonianCLdot \leq \twist\transpose\wrenchExt.\label{eq:passivityCondition}
\end{align}
Following similar steps as in~\cite{Rashad2019Tank}, the time derivative of $\hamiltonianCL$ results in
\begin{align}
\begin{split}
\hamiltonianCLdot =&-\underbrace{\twist\transpose\Kdampingbar \twist}_{\dampingone}
-\underbrace{\momentumEst\transpose\Kobserver\transpose\Kobserver\momentumEst}_{\dampingtwo}\\
&\quad+\twist\transpose\wrenchExt
+\underbrace{\twist\transpose}_{\effortA}\underbrace{\wrenchCmdTracking}_{\drainingA} + \underbrace{\twist\transpose}_{\effortBC}\underbrace{\Mmatrix\wrenchExtEst}_{\drainingBC}\\
&\quad+\underbrace{\momentumEst\transpose\Kobserver\transpose}_{\effortDE}\underbrace{\left(\Cmatrix\twist+\wrenchGravity+\wrenchCommand+\Kobserver\momentum\right)}_{\drainingDE\coloneqq\momentumEstDot+\Kobserver\momentumEst},
\end{split}\label{eq:hcl}
\end{align}
where $p_i\coloneqq\bm{y}_i\transpose\bm{\omega}_i,\ i\in\{1,2,3\}$ 
are the potentially passivity-violating power flows, since their signs are a priori undefined. 

Analyzing \cref{eq:hcl}, we can identify two damping terms, $\dampingone$ and $\dampingtwo$ which represent energy flowing out of the system. Conversely, the last three terms in \cref{eq:hcl} can be positive. As \cref{eq:passivityCondition} can be violated, passivity cannot be always guaranteed.
In order to ensure an overall passivity of the closed loop system, we extend the system dynamics with a virtual energy tank presented in the next section. 

\section{ENERGY TANK}

This virtual tank works as a energy reservoir that can be filled with the energy dissipated by damping terms and drained by the terms that can add energy to the system to compensate for them in the total energy balance. By designing the tank in- and outflows properly, we can achieve a cancellation of the passivity-violating terms in the total closed-loop dynamics.
We analyze \cref{eq:hcl} to identify the following power flows that can lead to a non-passive system:
\begin{itemize}
\item $\powerA=\twist\transpose\wrenchCmdTracking$, coming from wrench tracking commands;
\item $\powerBC=\twist\transpose\Mmatrix\wrenchExtEst$, coming from components of the impedance controller;
\item $\powerDE=\momentumEst\transpose\Kobserver\transpose\drainingDE$, coming from the wrench observer.
\end{itemize}
We then define the energy tank state as $\tankState(t)\in\nR{}$ and the tank energy as $\tankEnergy=\frac{1}{2}\tankState^2$. In order to constrain the available energy in the tank, we introduce a lower and an upper bound, $\tankLowerLimit$ and $\tankUpperLimit$, respectively, such that $0 < \tankLowerLimit < \tankUpperLimit$.

We define the state time derivative as
\begin{align}
\tankStatedot=\frac{\beta}{\tankState}\left(\eta_1\dampingone+\eta_2\dampingtwo\right)+\tankInput,
\end{align}
where $\tankInput$ represents the tank input and $\beta\in (0,1)$ prevents the tank energy to exceed the maximum value. In particular, $\beta=1$ if $E_t\leq\tankUpperLimit$ and $\beta=0$ otherwise.
We can use $\eta_1, \eta_2\in\left[0,1\right]$ to control how much power is fed from the system to the tank.
If there is enough energy available. the tank input is defined such that the tank reserve of energy can be used to compensate actions violating passivity.
In particular,
\begin{align}
\tankInput(t)=-\frac{1}{\tankState}\sum_{i=1}^3 p_i.
\end{align}
We ensure that $\tankState>0\ \forall\ t$ by adding a lower-limit policy in \cref{sec:lowerlimit}.

\subsection{Passivity analysis}
The tank energy dynamics are then given by
\begin{align}
\begin{split}
\hamiltonianTankdot&=\tankState\tankStatedot
=\beta\left(\eta_1\dampingone+\eta_2\dampingtwo\right)-\sum_{i=1}^3 p_i.
\end{split}
\end{align}
Considering the tank together within the closed loop dynamics, we can write the energy dynamics of the entire system:
\begin{align}
\begin{split}
\hamiltonianCLtotaldot&=\hamiltonianCLdot+\hamiltonianTankdot=-\dampingone-\dampingtwo+\twist\transpose\wrenchExt+\sum_{i=1}^3 p_i + \hamiltonianTankdot\\
&=-(1-\beta\eta_1)\dampingone-(1-\beta\eta_2)\dampingtwo+\twist\transpose\wrenchExt\leq\twist\transpose\wrenchExt.
\end{split}\label{eq:passivity2}
\end{align}
Thanks to the introduction of the tank, the passivity of the system can be ensured as long as the tank is not fully drained.

\section{POWER VALVES AND SAFETY POLICIES}\label{sec:policies}

From the previous section it appears that the system passivity, as so stability, can be preserved for every control action as long as the energy tank is not drained to the minimum. 
Therefore, it is important to design a method to limit the draining of the tank, reducing the non-passive actions and eventually set them to zero when the tank energy reaches its minimum value.
This can be done employing the concept to \textit{power valves} introduced in~\cite{Shahriari2018} together with some safety policies to regulate their values.

\subsection{Power valves}
In order to not only maintain passivity but also in an effort to maximize the chances of executing the given task, we use power valves to control the power flows of passivity-violating tasks individually.
We would therefore like to scale the power flows that can lead to a non-passive system such that
\begin{align}
p^*_i=\bm{y}_i\transpose\bm{\Gamma}_i \bm{\omega}_i.
\end{align}
where $\bm{\Gamma}_i=\diag{\gamma_{i,1},\dots,\gamma_{i,6}},i\in\{1,2,3\}$ is a tuning parameter described in the following.
This is equivalent to imposing:
\begin{subequations}
\begin{align}
\drainingA^*&=\wrenchCmdTracking^*= \GammaA\wrenchCmdTracking\\
\drainingBC^*& =\GammaBC\Mmatrix\wrenchExtEst\\
\begin{split}
\drainingDE^*&=\GammaDE\drainingDE=\momentumEstDot^*+\Kobserver\momentumEst^*.
\end{split}
\end{align}\label{eq:scaledCommands}
\end{subequations}
It is easy to verify that the new scaled power flows can be obtained by modifying the controller and wrench observer as follows:
\begin{align}
\begin{split}
\wrenchCmd^*&=\wrenchCmdTracking^*+\wrenchCmdImpedance^*\\
&=\GammaA\wrenchCmdTracking+\GammaBC\Mmatrix\wrenchExtEst\\ &\qquad\qquad\qquad-\left(\Kdampingbar\twist+\Kspringbar\error\right)-\Cmatrix\twist-\wrenchGravity
\end{split}\\
\momentumEstDot^*&=\GammaDE\drainingDE-\Kobserver\momentumEst^*.
\end{align}

\subsection{Valve gain scaling}
As shown in \cref{eq:passivity2} the system is passive for any $\bm{\Gamma}_i$. However, the exact values of $\bm{\Gamma}_i$ determine the performance in executing the three passivity-violating tasks of external wrench estimation, impedance control, and wrench tracking control. \replaced{We present different approaches in order to adaptively set the valve gains}{We present different approach setup those valve gains} according to the following goals (in decreasing order of priority):
\begin{enumerate}
\item Safe interaction;
\item Correct estimation of external wrenches;
\item Impedance control;
\item Tracking of a reference force.
\end{enumerate}
These goals can be translated into the following valve policies:
\begin{enumerate}
\item Limit the total power flow from the tank to the system, i.e., $-\hamiltonianTankdot\leq\powerMax$;\label{list:a}
\item Maximize $p_3^*$, obtained by maximizing $\GammaDE$;
\item Maximize $p_2^*$, obtained by maximizing $\GammaBC$;
\item Maximize $p_1^*$, obtained by maximizing $\GammaA$.
\end{enumerate}

In the following we present three different policies to achieve these goals. 
\paragraph{Individual Gain Scaling (IGS)}
The valve gains are scaled according to individually set maximum power limits:
\begin{align}
\gamma_i = \begin{cases}\frac{p^+_i}{p_i} \quad &\text{if }p_i>p^+_i,\\
1 &\text{else.}\end{cases}
\end{align}
\paragraph{Weighted Gain Scaling (WGS)}
The gains are scaled such that the total power outflow is limited to $p_{tot}^+$ and the ratios of individual power flows are determined by $\delta_i$:
\begin{align}
\gamma_i = \begin{cases}\frac{\delta_i \powerMax}{\sum \delta_i p_i} \quad &\text{if }\sum p_i>p_{tot}^+,\\
1 &\text{else.}\end{cases}
\end{align}

\paragraph{Sequential Gain Assignment (SGA)}
Tasks of highest priority are fully admitted until a maximum power limit is reached. The last valve to be actived is scaled so that the maximum allowable power flow is achieved. If the valves are sorted by priorities in descending order (i.e. lower index represents higher priority), then:
\begin{align}
\gamma_i = \begin{cases}
1 \quad &\text{if } \sum_{j=1}^i p_j\leq p_{tot}^+,\\
\frac{p_{tot}^+-\sum_{j=1}^{i-1}p_j}{p_i} &\text{if }\sum_{j=1}^i p_j\geq p_{tot}^+,\\
0 &\text{else.}\end{cases} 
\end{align}

\subsection{Lower tank limits}\label{sec:lowerlimit}
In addition to the individual valves, we add a general multiplier $\alpha$ to prevent the tank from draining too close to a lower limit $\tankLowerLimit$.  This gain is multiplied with all valve gains, such that $\bm{\Gamma}_i^*=\alpha\cdot\bm{\Gamma}_i$.
In order to achieve a smooth scaling close to this lower bound we employ a cosine step function between $\tankLowerLimit$ and a threshold $\tankEps$:

\begin{align}
\alpha=\begin{cases}
0 & \text{if }\tankEnergy\leq\tankLowerLimit,\\\
\frac{1}{2}\left(1-\cos\left(\frac{\tankEnergy-\tankLowerLimit}{\tankEps}\pi\right)\right) & \text{if }\tankLowerLimit\leq\tankEnergy\leq\tankLowerLimit+\tankEps,\\
1 & \text{else.}
\end{cases}
\end{align}

\section{EXPERIMENTAL RESULTS}\label{sec:experiments}

\subsection{Experimental setup}
For experiments we use the OMAV presented in~\cite{Bodie2019} and shown in \cref{fig:frames}.
It is designed with six equally spaced arms with double rotor groups, which can be tilted by servo-motors obtaining omnidirectional thrust vectoring.
A rigid arm is attached to the body, pointing along its positive $\vec{x}_B$-axis.
Sensor fusion of an Inertial Measurement Unit (IMU) with an external motion capture system is used for state estimation.
A \ac{FT} sensor measures the ground truth force acting at the end-effector. Its measurements $\vec{f}_{meas}$ are used for comparisons only.

\subsection{Controller implementation}
The implementation of the valve policies in a discrete-time controller can lead to strong chattering behavior of the valve gains and the respective power flows.
As the valve gains are computed based on the power flows at the previous time step, this can result in alternating exceedingly high and low power flows.
In order to mitigate this phenomenon, we apply a first order lowpass filter on $\bm{\Gamma}_i$ with a cutoff frequency of \SI{2}{\hertz}. 
While this does not violate passivity, it can lead to an unnecessarily conservative behavior where the control objectives are not fully satisfied because the valves are slowly restored to high values after low ones.  %
Furthermore, we simplify the valve gain matrices by using five scalars $\gamma_i$, resulting in the following multipliers:
\begin{align}
\begin{split}
\GammaA&=\diag{\gamma_1, \zeros{1\times 5}}\\
\GammaBC&=\blkdiag{\valvesB\eye{3}, \valvesC\eye{3}}\\
\GammaDE&=\blkdiag{\valvesD\eye{3}, \valvesE\eye{3}}.
\end{split}
\end{align}
The choice of $\GammaA$ leads to a force tracking controller along the body $x$-axis only, ignoring wrench errors in other directions, which is the goal of the task. We further separate $\GammaBC$ and $\GammaDE$ into linear and angular components as their respective power flows have naturally different magnitudes.

\subsection{Comparison of different policies}
We evaluate the different policies by pushing with the end-effector against a cart which is able to move freely on a flat ground (see \cref{fig:frames}). This is done by first approaching the cart and then activating the \ac{WTC}.
\added{Once the WTC is activated, the impedance controller remains enabled to track a constant position and attitude.}

For all experiments we use $\eta_1=\eta_2=0.4$ to ensure a high passivity margin, since values around $\eta_i=1$ can lead to the power in- and outflows balancing each other out.

\subsubsection*{Individual Gain Scaling}
We tuned the five power limits to the following values: $p^+_1=1.0$, $p^+_{2,lin}=1.0$, $p^+_{2,ang}=1.0$, $p^+_{3,lin}=30.0$, $p^+_{3,ang}=3.0$ [\si{\watt}]. 
This configuration was chosen following this reasoning: The correct estimation of the external wrench is of highest priority, resulting in the highest allowable power flows in $p_3$. However, as the power flows created by the linear dynamics are generally about one magnitude larger than the ones created by angular dynamics (given our platform design and dynamic capabilities), we allow $p_{3,lin}$ to be larger than $p_{3,ang}$. The force tracking and impedance power flows are limited to lower values as they can otherwise lead to dangerously large forces.

\begin{figure}[t]
    \centering
    \vspace*{-0.2cm}
    \includegraphics[width=\linewidth]{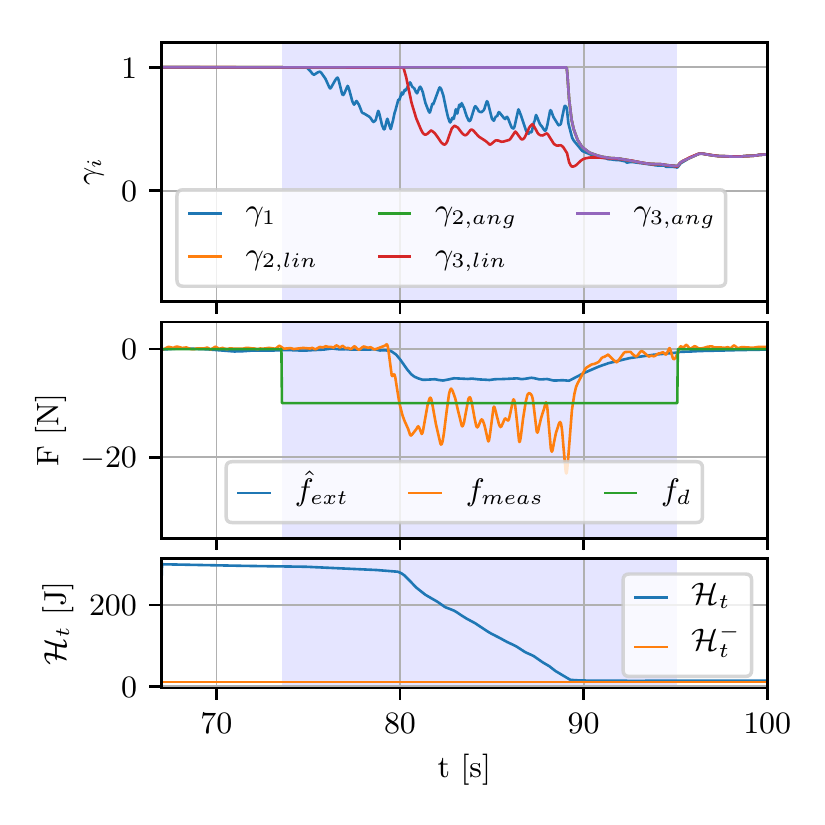}
    \vspace*{-0.7cm}
    \caption{Valve states, interaction forces, and tank energy for IGS. $\hat{f}_{ext}$ is the estimated interaction force, $f_{meas}$ the \ac{FT} sensor measurement, and $f_d$ the reference force.  The blue shaded area highlights the period during which the WTC is enabled.}
    \vspace*{-0.2cm}
    \label{fig:valve_states_igs}
\end{figure}

\Cref{fig:valve_states_igs} shows the valve gains and tank energy during one push experiment. Even though the OMAV was at a distance of \SI{0.5}{\meter} upon activation of the WTC, the approach to the (a priori unknown) contact point was performed smoothly due to the wrench tracking power reduction through $\gamma_1$.
After contact at $t=\SI{80}{\second}$,  the tank drains quickly through the wrench observer port, leading to a complete draining at which the tracking command is ramped down, stopping the movement. The cart traveled a distance of \SI{0.3}{\meter} in this period. Note that the estimated and measured interaction force differ due to the limited power flow $p_3$ of the observer. 

\subsubsection*{Weighted Gain Scaling}
\begin{figure}[t]
    \centering
    \vspace*{-0.2cm}
    \includegraphics[width=\linewidth]{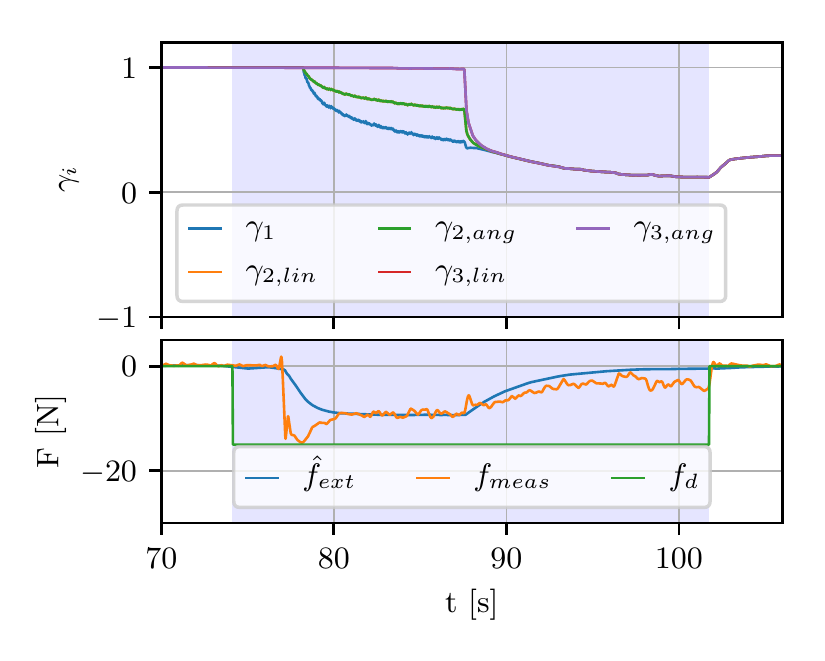}
    \vspace*{-0.7cm}
    \caption{Valve states and force tracking for WGS. The valves are scaled smoother than in IGS and the interaction force is estimated correctly.}
    \vspace*{-0.4cm}
    \label{fig:valve_states_wgs}
\end{figure}

For WGS we used a maximum allowed power output of $p_{tot}^+=\SI{90}{\newton}$ and the following weighting:
$\delta_1=1.0$, $\delta_2=5.0$, $\delta_3=5.0$, $\delta_4=10.0$, $\delta_5=10.0$. 
Again we assign the highest priorities to the correct wrench estimation and the lowest priority to tracking the reference force. 
This is reflected in the experimental results shown in \cref{fig:valve_states_wgs}. 
After establishing contact with the cart, the force tracking valve is decreased together with the impedance torque component, until the tank is drained completely and tracking is stopped.
The interaction force estimate is more accurate compared to IGS, also resulting in a smoother force tracking.
While the valve-scaling behavior of this approach turns out to be not only smoother than in IGS, the tuning of a single maximum power output rather than many individual limits is also more convenient in practice.

\subsubsection*{Sequential Gain Assignment}
The priorities for SGA were set in the following order (from high to low priority): $\gamma_5$, $\gamma_4$, $\gamma_3$, $\gamma_2$, $\gamma_1$.  
As the experimental result was similar to the result of WGS, we do not show the related plots for the lack of space.
However, this method proved to be very practical, as it depends --- once the priority order has been determined --- on only one tuning variable (i.e. $p_{tot}^+$). That way, an interaction task can be attempted with initially low values of $p_{tot}^+$ in order to guarantee a safe behavior. If the limit is too low for task execution, it can gradually be increased, until the task can be executed.

\subsubsection*{Discussion} As the three policies have been tuned with the same objectives in mind, the experimental results are similar. However, the tuning procedures are different and turned out to be less tedious for WGS and SGA as compared to IGS.

\subsection{Comparison of IGS and no policy}
\begin{figure}[h]
    \centering
    \vspace*{-0.1cm}
    \includegraphics[width=\linewidth]{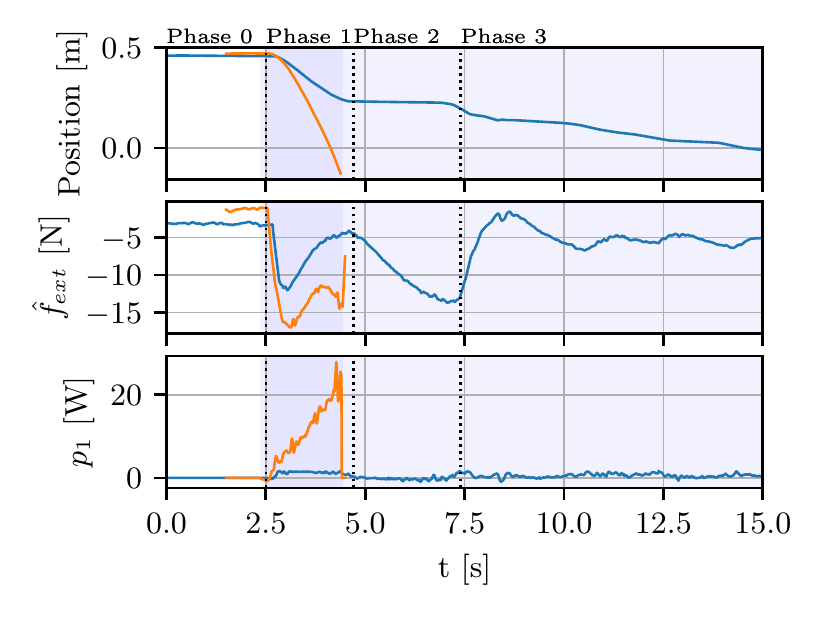}
    \vspace*{-0.6cm}
    \caption{Pushing a cart over an obstacle. The WTC is activated at $t=\SI{2.5}{\second}$. The blue lines represent the outcome with using IGS, the orange line without policies.}
    \vspace*{-0.2cm}
    \label{fig:comparison_cart_pushing}
\end{figure}

Lastly, we compare the experiment of pushing a cart over a small obstacle on the ground with employing IGS against employing no policy.
For this experiment we assume only a minimal knowledge about the task requirements, which is that a force of $\SI{15}{\newton}$ is sufficient to overcome the obstacle. Other parameters such as the state, mass, or friction of the cart are unknown.
The maximum allowed power value for the wrench tracking command is set to $p_1^+=\SI{1.5}{\watt}$. The results are presented in \cref{fig:comparison_cart_pushing}. Both IGS and no policy are shown, highlighting that the former is able to perform the task while the latter had to be stopped after accelerating too quickly. The different phases are described in the following.

At the start of the experiment, the OMAV is hovering with the end-effector touching the cart (phase 0).
In phase 1, the \ac{WTC} is activated with a constant reference interaction force of $f_{ref}=\SI{16}{\newton}$.  In phase 2, the cart hits the obstacle and a higher force is needed to overcome it.
Once the obstacle has been passed (phase 3), the cart is pushed with limited power, maintaining stability.
Eventually, the reference interaction force is never reached due to the power policy.
As opposed to IGS, not using a policy leads to a fast increase of $\wrenchCmdTracking$ and consequently a high acceleration of the platform, from which it eventually becomes unstable.

Employing valve gain scaling based on power flows allows to push an unmodeled movable cart, resulting in a safe and robust interaction. 
Thanks to the formal proof of passivity, the interaction with any passive environment is guaranteed to be stable, i.e., also in the case of a cart with different mass or friction parameters.
Additionally, we found that scaling power flows based on their magnitudes rather than on the tank energy is especially suitable for highly dynamic systems such as \acp{OMAV}. 
In fact, we experimentally experienced that the system might diverge when too high power flows limits are set. 
Limiting these power flows adds another safety layer to the concept of stopping interactions upon a drained tank.

\section{CONCLUSIONS}\label{sec:conclusions}
We applied energy tank-based policies for robust interaction with moving environments, where no knowledge nor assumption regarding the environment is required.
We observed that especially in interaction with moving objects, the power policies prevent the tank from providing harmful amounts of energy within a short time by limiting the interaction power.
The policies are able to reduce the performance of the corresponding control action and ensure a safe dynamic interaction.
Scaling the valves according to the described policies proved to be an efficient method to keep power flows within bounds. 
\added{While the approach ensures safe interaction, it can result in overly conservative control actions which can impede the execution of a task. This could be addressed by adapting the policies online, which we leave to future research.}

\bibliographystyle{IEEEtran}
\bibliography{./bibAlias,bib}

\end{document}